\documentclass[runningheads]{llncs}
\usepackage[T1]{fontenc}
\usepackage{amsmath}
\usepackage{amssymb}
\usepackage{graphicx}
\usepackage{multirow}
\usepackage{booktabs}
\usepackage{caption}
\usepackage{subcaption}
\usepackage{dirtytalk}
\usepackage{xspace}
\usepackage{arydshln}
\usepackage{bm}

\DeclareMathOperator{\argmin}{arg\,min}
\DeclareMathOperator{\argmax}{arg\,max}

\newcommand{\seg}{segmentation\xspace}
\newcommand{\segs}{segmentations\xspace}
\newcommand{\methodname}{StyleSeg\xspace}
\newcommand{\ssegname}{\texttt{SSeg}\xspace}
\newcommand{\msegname}{\texttt{MSeg}\xspace}
\newcommand{\groundtruth}{ground truth\xspace}
\newcommand{\dermofit}{DermoFit\xspace}
\newcommand{\phtwo}{PH\textsuperscript{2}\xspace}
\newcommand{\suppmat}{Supp. Mat.\xspace}
\newcommand{\isictestname}{ISIC Archive-Test\xspace}
\newcommand{\newdatasetname}{ISIC-MultiAnnot\xspace}
\newcommand{\newmetricname}{Annotator-Style Alignment Strength\xspace}
\newcommand{\newmetricnameshort}{AS\textsuperscript{2}\xspace}
\newcommand{\nstyles}{M}
\newcommand{\classmodel}{f_c}
\newcommand{\segmodel}{f_s}
\newcommand{\classmodelparams}{\mathrm{\Theta}_c}
\newcommand{\segmodelparams}{\mathrm{\Theta}_s}
\usepackage[colorlinks]{hyperref}
\newcommand{\ghrepo}{https://github.com/sfu-mial/StyleSeg}

\usepackage{color}

\newcommand{\camready}[1]{{#1}}
\newcommand{\beststyle}{\mathcal{J}}

\begin{document}
\title{
Segmentation Style Discovery:\\Application to Skin Lesion Images
% \thanks{Supported by organization x.}
}
\titlerunning{Segmentation Style Discovery}
\author{
Kumar Abhishek\inst{1}\orcidID{0000-0002-7341-9617} \and
Jeremy Kawahara\inst{2}\orcidID{0000-0002-6406-5300} \and
Ghassan Hamarneh\inst{1}\orcidID{0000-0001-5040-7448}
}
% index{Abhishek, Kumar}
% index{Kawahara, Jeremy}
% index{Hamarneh, Ghassan}
%
\authorrunning{
Abhishek et al.
}
\institute{
School of Computing Science, Simon Fraser University, Canada \and
AIP Labs, Budapest, Hungary\\
\email{\{kabhishe,hamarneh\}@sfu.ca, jeremy@aip.ai}
}
\maketitle              % typeset the header of the contribution
\begin{abstract}

Variability in medical image segmentation, arising from annotator preferences, expertise, and their choice of tools, has been well documented. While the majority of multi-annotator segmentation approaches focus on modeling annotator-specific preferences, they require annotator-segmentation correspondence. In this work, we introduce the problem of segmentation style discovery, and propose \methodname, a \seg method that learns plausible, diverse, and semantically consistent \seg styles from a corpus of image-mask pairs without any knowledge of annotator correspondence. \methodname consistently outperforms competing methods on four publicly available skin lesion segmentation (SLS) datasets. We also curate 
\newdatasetname, the largest multi-annotator SLS dataset with annotator correspondence, and our results show a strong alignment, using our newly proposed measure \newmetricnameshort, between the predicted styles and annotator preferences.
The code and the dataset are available at \url{\ghrepo}.

\end{abstract}
\section{Introduction}
\label{sec:introduction}

Medical image \seg is a critical component in medical image analysis pipelines, either as a preprocessing step for subsequent analyses or for treatment planning and image-guided human or robotic intervention. 
Following the seminal works of Long et al.~\cite{long2015fully} and Ronneberger et al.~\cite{ronneberger2015u}, there has been tremendous progress in deep learning (DL)-based medical image \seg~\cite{asgari2021deep}. The majority of these works focus on learning to predict a single \seg for an image. However, variability among experts when segmenting images has been well-documented, and these resulting \seg masks are the product of latent factors such as ambiguous object boundaries and differences in tools, annotators' skill levels, criteria, and approaches to \seg, and they capture different annotator \seg preferences or \say{styles}. Without accommodating these variations, a \seg model optimized to minimize training error over a variety of human annotations may produce an \say{average} \seg. This has motivated research that can be broadly categorized into two classes: methods that model and learn to predict a single \say{gold standard} segmentation through label aggregation~\cite{warfield2004simultaneous,kats2019soft,mirikharaji2021d} (\textbf{\ssegname}) and methods that predict multiple \segs to capture the variability of annotations~\cite{rupprecht2017learning,zhang2020learning,ji2021learning} (\textbf{\msegname}).

\textbf{\msegname} methods rely on modeling annotator-specific preferences, and training them typically requires annotations with annotator-\seg correspondence. Therefore, given a set of images and a set of annotators, annotator-\seg correspondences can be represented as a bipartite graph  when every image has been segmented by at least 1 annotator, e.g., LIDC-IDRI~\cite{armato2011lung}, or a complete bipartite graph when every image has been segmented by every annotator, e.g., RIGA~\cite{almazroa2017agreement}. 
However, in the absence of such a correspondence, i.e., a scenario where we have a corpus of images and corresponding annotations without any knowledge of annotator IDs, defining a segmentation style is non-trivial since the latent factors associated with each \seg are unknown, thus making it challenging to explicitly train a segmentation model to reproduce a particular style.
Since we are unable to confirm even the number of unique annotators, we hypothesize that a possible solution for modeling multi-annotator segmentations would be \textbf{discovering unique annotation styles} from the dataset alone. 
Such a discovery-based approach needs to ensure (1) diversity in the discovered styles, (2) \seg plausibility across all the styles, and (3) semantic consistency of the \segs across all the images. 
However, to the best of our knowledge, there is minimal prior work on the discovery and modeling of annotation styles in the absence of annotator correspondence.

We argue that since even experts can (considerably) differ in 
how they segment,
it is only natural that automated models trained thereupon also exhibit this variety. We envision that a \seg system should produce results that align with the expectations of its (clinical) users, and that these users can vary in their personalization preferences (e.g., a study~\cite{fortina2012where} found that expert dermatologists prefer \say{tighter} \segs than their inexperienced counterparts). Moreover, such a system should, with minimal supervision, continue to produce the style that a user expects, thus avoiding constant user involvement with either manual corrections or image-by-image selection of preferred \seg style.

In this work, we tackle the problem of style discovery and personalization modeling in medical image \seg without requiring annotator correspondence, and focus our analysis on skin lesion \seg (SLS). Advancements in DL over the past decade as well as the availability of large publicly available annotated datasets have enabled large strides in SLS~\cite{mirikharaji2023survey,tschandl2019domain}. Therefore, in this work, we work on style discovery in the context of multiple annotators for SLS, which has
not been explored extensively. The majority of previous works focus on \textbf{\ssegname} methods: either to select training samples that have high inter-annotator agreement~\cite{ribeiro2020less} or training ensemble models to handle annotators' variability~\cite{mirikharaji2021d}. More recently, Zepf et al.~\cite{zepf2023label} presented a small-scale ($n=300$) analysis of annotation styles in images from the ISIC 2019 dataset based on the granularity of the annotation boundaries. In this work, we make the following contributions: 
\textbf{(1)} we introduce the problem of \seg style discovery in the absence of any annotator correspondence and propose a method (\methodname) that predicts multiple plausible, diverse, and semantically consistent \seg styles, \textbf{(2)} we curate,
to the best of our knowledge, the largest multi-annotator SLS dataset (\newdatasetname) with annotator-segmentation mapping, and \textbf{(3)} we introduce a new measure (\newmetricnameshort) for measuring the strength of alignment of the predicted styles with annotator preferences.

\section{Method}
\label{sec:method}

Let $\mathcal{X} = \{X_i\}_{i=1}^N$ be a set of images
and
corresponding segmentation masks $\mathcal{Y} = \{\{Y_{ik}\}_{k=1}^{K_i}\}_{i=1}^{N}$, 
where $K_i > 0$ denotes the number of different ways $X_i$ was segmented, without any knowledge of annotator correspondence.
The goal is to discover unique annotation \say{styles} in this data $(\mathcal{X}, \mathcal{Y})$ such that, when given an image $X_{i}$, we predict $\{Y_{ij}\}_{j=1}^{M}$: ${M}$ unique \segs of $X_i$, that are diverse, plausible for $X_i$, and are of semantically consistent styles across all images.

\begin{figure*}[ht!]
    \centering
    \begin{subfigure}[t]{\textwidth}
        \centering
        \includegraphics[width=\textwidth]{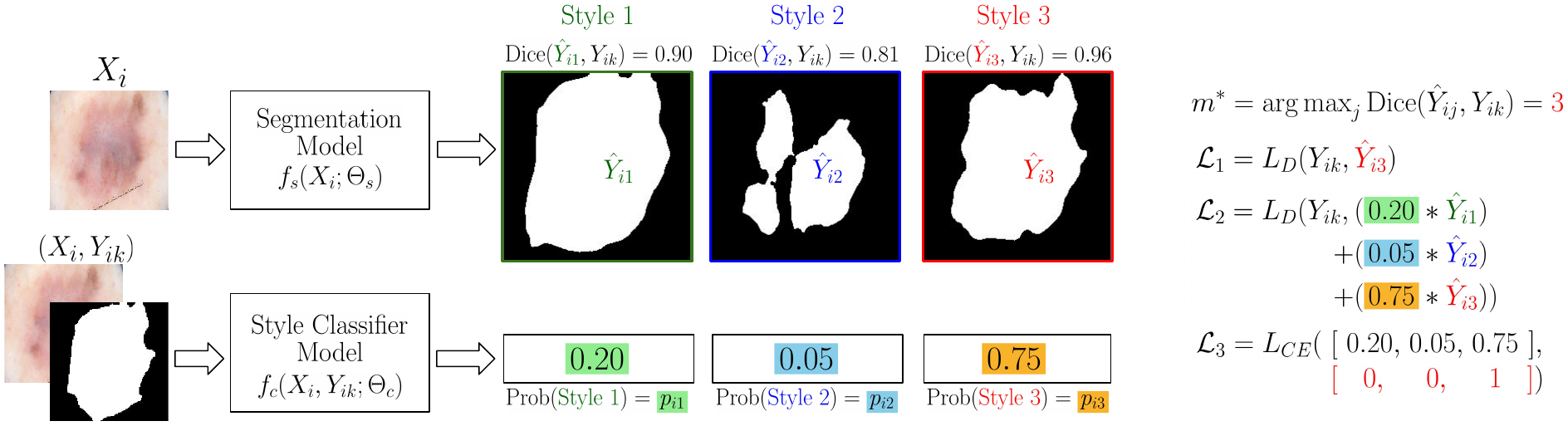}
        \caption{\methodname overview with $\nstyles = 3$: the \seg model $\segmodel$ predicts $3$ plausible \segs of different styles, while the style classifier $\classmodel$ predicts the style that is the most similar to the \groundtruth. 
        % $\segmodel, \classmodel$ are optimized by minimizing $\mathcal{L}_{\mathrm{total}} = \mathcal{L}_1 + \mathcal{L}_2 + \mathcal{L}_3.$
        }
    \end{subfigure}%
    \\
    \begin{subfigure}[t]{0.42\textwidth}
        \centering
        \includegraphics[width=\textwidth]{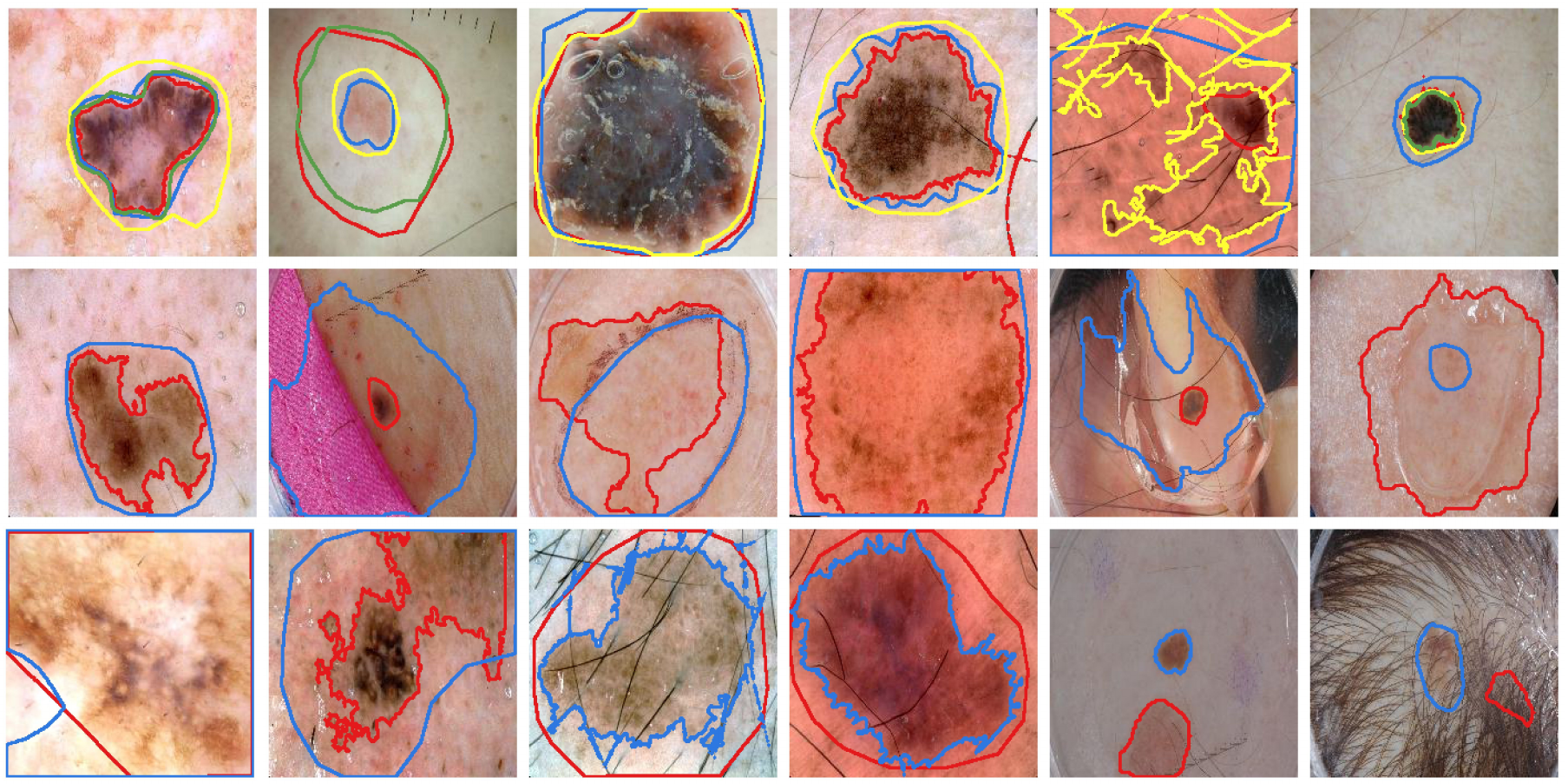}
        \caption{Sample training images showing variability of \segs.}
    \end{subfigure}
    \quad
    \begin{subfigure}[t]{0.53\textwidth}
        \centering
        \includegraphics[width=\textwidth]{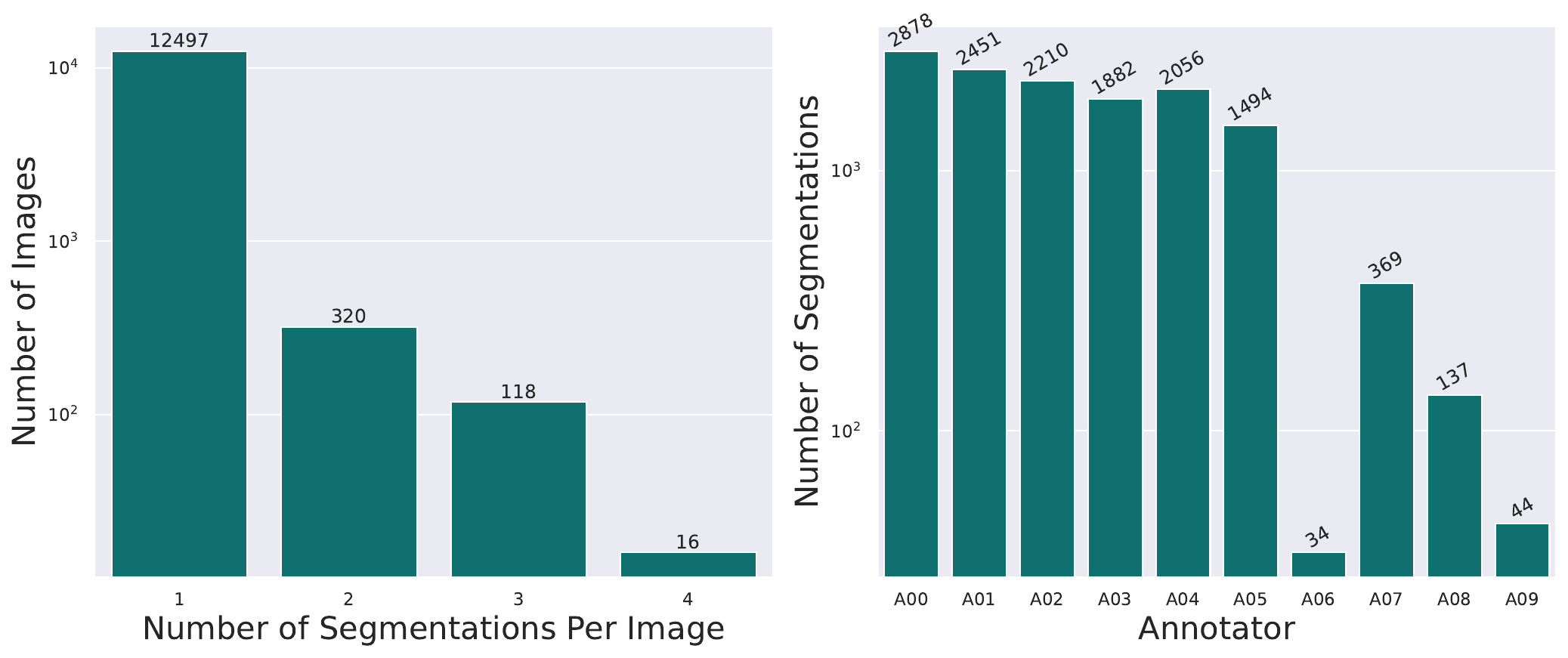}
        \caption{Distribution of \newdatasetname by number of annotators and segmentations.}
    \end{subfigure}
    \caption{(a) An overview of the proposed method \methodname. (b) Inter-annotator variability in the training images. (c) An annotator-wise breakdown of the newly curated \newdatasetname dataset.}
    \label{fig:datasets_overview}
\end{figure*}

To this end, we propose \methodname (Fig.~\ref{fig:datasets_overview} (a)): a segmentation approach that learns to predict $M$ plausible \segs that capture a variety of styles from a corpus of images and corresponding masks without any annotator correspondence. \methodname consists of two deep learning models that are trained together: (i) a \textbf{segmentation model} $\segmodel$, parameterized by $\segmodelparams$, that predicts $M$ segmentation masks from image $X_i$, \camready{where $M \in \mathbb{N}$ is a user-specified value,}
\begin{equation}
    \{\hat{Y}_{ij}\}_{j=1}^{M} = \segmodel(X_i; \segmodelparams),
\end{equation}
and (ii) a \textbf{style classifier model} $\classmodel$, parameterized by $\classmodelparams$, that predicts a vector $p_i \in \mathbb{R}^{M}$ of $M$ probabilities,
\begin{equation}
\label{eqn:classifier}
    p_i = \classmodel(X_i, Y_{ik}; \classmodelparams),
\end{equation}
where
$p_{ij}$ is the probability that $(X_i, Y_{ik})$ is of style $j$. 
Note that knowing $X_i$ is necessary to define the styles, since the observed \segs are a product of image content and annotation style. 

Of the $\nstyles$ predicted \segs from $\segmodel$, we first need to identify the style that is the closest to the ground truth $Y_{ik}$, and then optimize it to make it even closer. Mathematically, we minimize the loss $\mathcal{L}_1$, 
\begin{align}
   \mathcal{L}_1 &= L_D(Y_{ik}, \hat{Y}_{im^*}), \\
   m^* &= \argmax_j \mathrm{Dice}(Y_{ik}, \hat{Y}_{ij}), 
\end{align}
where $L_D = 1-\mathrm{Dice}$ and $\mathrm{Dice}$ denotes the Dice similarity coefficient~\cite{dice1945measures}.
We also require the other predicted styles to still be plausible, i.e., similar to \groundtruth $Y_{ik}$. However, requiring all styles to be equally plausible compromises the styles' diversity. 
Therefore, we make the strength of a style's plausibility requirement proportional to its likelihood of being the predicted style, according to the style classifier $\classmodel$ (Eqn.~\ref{eqn:classifier}).
To this end, we encourage the weighted sum of predicted \segs $\hat{Y}_{ij}$ to be similar to \groundtruth $Y_{ik}$, where the \camready{scalar} weights are $p_{ij}$. Mathematically, we minimize the loss $\mathcal{L}_2$,
\begin{equation}
\label{eqn:l2}
    \mathcal{L}_2 = L_D\left(Y_{ik}, \sum\nolimits_j^M p_{ij} \hat{Y}_{ij}\right).
\end{equation}

Weighting the $M$ \segs by $p_i$ ensures that when $p_i$ has a high entropy (e.g., in the initial training epochs), all styles are encouraged to be similar to $Y_{ik}$, whereas when $p_i$ has a low entropy, only a subset of the $M$ styles are encouraged to be similar to $Y_{ik}$, thus enabling a coarse-to-fine style refinement. 

Additionally, we employ a cross-entropy loss $\mathcal{L}_3$ to train the style classifier $\classmodel$ by learning to predict the style that is the most similar to the \groundtruth,
\begin{equation}
    \mathcal{L}_3 = L_{CE}(p_i, m^*).
\end{equation}

Finally, we optimize the parameters $\segmodelparams$ of $ \segmodel$ and $\classmodelparams$ of $\classmodel$ using 
\begin{equation}
    \segmodelparams^*, \classmodelparams^* = \argmin_{\segmodelparams, \classmodelparams} \sum\nolimits_i^N \mathcal{L}_{\mathrm{total}},
\end{equation}
\noindent where
\begin{equation}
\label{eqn:total_loss}
    \mathcal{L}_{\mathrm{total}} = \mathcal{L}_1 + \mathcal{L}_2 + \mathcal{L}_3.
\end{equation}
Note that we do not include an explicit style distinctiveness constraint since, in the absence of annotator correspondence, the styles are entangled with the \segs. Nevertheless, these loss terms %$L_{CE}$ and $D_1(\cdot)$ 
used together (Eqn.~\ref{eqn:total_loss}) implicitly encourage the styles to be different as the training progresses, as seen in our results.

\begin{figure*}[ht!]
    \centering
    \begin{subfigure}[t]{0.49\textwidth}
        \centering
        \includegraphics[width=\textwidth]{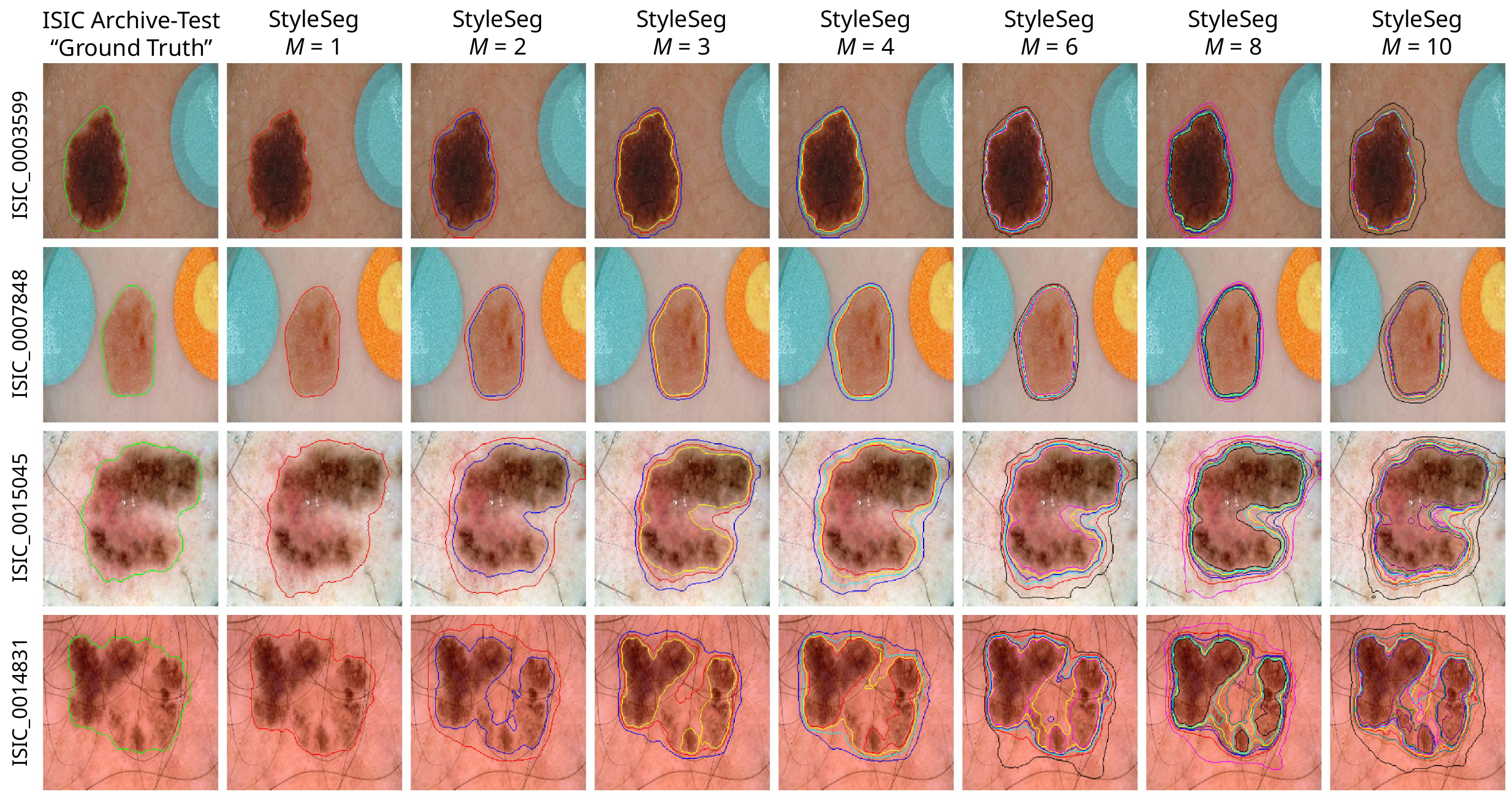}
        \caption{Sample outputs of \methodname for lesions with and without distinct borders.}
    \end{subfigure}%
    \hfill
    \begin{subfigure}[t]{0.49\textwidth}
        \centering
        \includegraphics[width=\textwidth]{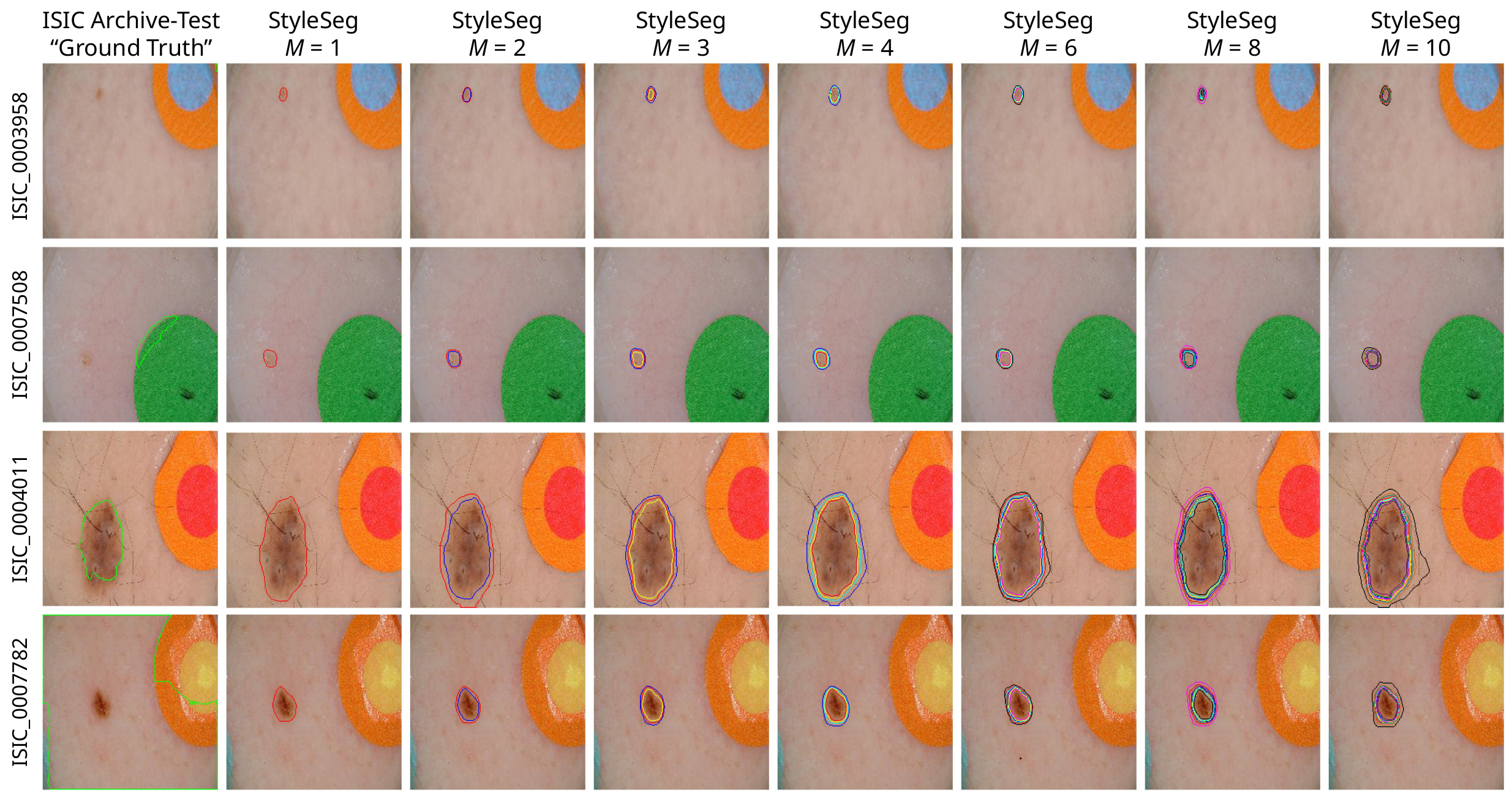}
        \caption{\methodname produces better \segs than even the test \say{\groundtruth{}}.}
    \end{subfigure}
    \caption{Evaluating \methodname on \isictestname: diverse and plausible \segs that are semantically consistent across styles.}
    \label{fig:isic_v1_outputs}
\end{figure*}

\section{Results and Discussion}
\label{sec:results}

\underline{\textbf{Datasets:}} Similar to previous works~\cite{ribeiro2020less,mirikharaji2021d}, we train \methodname on images obtained from the ISIC Archive~\cite{ISICArchive}, specifically images with more than one \say{\groundtruth{}} \seg. We select 2,261 images that meet this criterion (2,122 with two, 100 with three, 35 with four, and 4 with five \segs), resulting in 4,704 image-mask pairs. 
Note that these images exhibit a vast range of inter-annotator agreement, as evidenced qualitatively (sample images with their masks in Fig.~\ref{fig:datasets_overview} (c)) and quantitatively (pairwise Dice coefficients and Fleiss' kappa in \suppmat Fig.~\ref{fig:suppmat_fig1} (a)). 
We choose Fleiss' kappa~\cite{fleiss1971measuring} over Cohen's kappa~\cite{cohen1960coefficient} used by Ribeiro et al.~\cite{ribeiro2020less} because the former can be used with multi-rater settings while the latter cannot~\cite{powers2012problem}. We reserve 1,525 image-mask pairs from the ISIC Archive for our validation set. See \suppmat for model architectures and training details.
We evaluate on
four publicly available datasets: \isictestname containing 10,000 dermoscopic images with just one \seg \groundtruth per image from the ISIC Archive, \dermofit (1,300 clinical images)~\cite{ballerini2013color}, SCD (206 dermoscopic images)~\cite{glaister2013msim}, and \phtwo (200 dermoscopic images)~\cite{mendonca2013ph}.

\noindent \underline{\textbf{Competing methods:}} We train \methodname with $M = \{2, 3, 4, 6, 8, 10\}$ and compare it to the following \ssegname methods: \texttt{NaiveTraining}: a \seg model without any annotator-specific knowledge; \texttt{RandAnnotID}~\cite{mirikharaji2021d}: 4 \seg models, one optimized for each annotator randomly assigned to a mask, \texttt{LessIs\-More}~\cite{ribeiro2020less}: a \seg model trained on a subset of the masks whose average pairwise Cohen's kappa is above 0.5; and \texttt{D-LEMA}~\cite{mirikharaji2021d}: an ensemble of Bayesian \seg models.
We also compare 
against an \msegname method \texttt{MHP} (multiple hypothesis prediction)~\cite{rupprecht2017learning} also with $\nstyles = \{2,3,4,6,8,10\}$.

\begin{table}[ht!]
\centering
\caption{Dice mean\tiny{\emph{std.dev.}} \normalsize comparing \methodname to \ssegname~\cite{ribeiro2020less,mirikharaji2021d} (first 4 rows) and \msegname (MHP~\cite{rupprecht2017learning}) methods. For the latter, we report the {mean, median, min., max.} of Dice between the ground truth and all the predicted \seg styles. Note how \methodname consistently outperforms all competing methods while also producing more plausible \segs than MHP.
% , as evidenced by higher minimum Dice for all values of $\nstyles$. 
$\oslash$ denotes that the result cannot be reported since D-LEMA's~\cite{mirikharaji2021d} code is not available.}
\label{tab:v1_results}
\resizebox{\textwidth}{!}{%
\setlength{\tabcolsep}{0.7em}
\def\arraystretch{1.25}
\begin{tabular}{@{}ccccc|cccc|cccc|cccc@{}}
\toprule
\multirow{2}{*}{Method} & \multicolumn{4}{c}{\isictestname ($n=10000$)}                                                                          & \multicolumn{4}{c}{PH\textsuperscript{2} ($n=200$)}                                                                                    & \multicolumn{4}{c}{DermoFit ($n=1300$)}                                                                              & \multicolumn{4}{c}{SCD ($n=206$)}                                                                                   \\ % \cmidrule(l){2-17} 
                  & Mean                     & Median                   & Minimum                  & Maximum                  & Mean                     & Median                   & Minimum                  & Maximum                   & Mean                     & Median                   & Minimum                  & Maximum                  & Mean                     & Median                   & Minimum                  & Maximum                  \\ \cmidrule(r){1-17}
NaiveTraining     & -                        & -                        & -                        & 0.800\tiny{\emph{0.188}} & -                        & -                        & -                        & 0.880\tiny{\emph{0.071}}   & -                        & -                        & -                        & 0.842\tiny{\emph{0.139}} & -                        & -                        & -                        & 0.766\tiny{\emph{0.198}} \\
RandAnnotID~\cite{mirikharaji2021d} & -                        & -                        & -                        & $\oslash$                        & -                        & -                        & -                        & 0.897\tiny{\emph{0.005}} & -                        & -                        & -                        & 0.826\tiny{\emph{0.004}} & -                        & -                        & -                        & $\oslash$                        \\
LessIsMore~\cite{ribeiro2020less}        & -                        & -                        & -                        & 0.815\tiny{\emph{0.178}} & -                        & -                        & -                        & 0.895\tiny{\emph{0.070}}  & -                        & -                        & -                        & 0.854\tiny{\emph{0.127}} & -                        & -                        & -                        & 0.804\tiny{\emph{0.169}} \\
D-LEMA~\cite{mirikharaji2021d}            & -                        & -                        & -                        & $\oslash$                        & -                        & -                        & -                        & 0.920\tiny{\emph{0.004}}  & -                        & -                        & -                        & 0.853\tiny{\emph{0.003}} & -                        & -                        & -                        & $\oslash$                        \\ \hdashline
2-MHP             & 0.796\tiny{\emph{0.168}} & 0.727\tiny{\emph{0.195}} & 0.727\tiny{\emph{0.195}} & 0.864\tiny{\emph{0.158}} & 0.850\tiny{\emph{0.114}} & 0.786\tiny{\emph{0.163}} & 0.786\tiny{\emph{0.163}} & 0.914\tiny{\emph{0.073}}  & 0.795\tiny{\emph{0.149}} & 0.707\tiny{\emph{0.229}} & 0.707\tiny{\emph{0.229}} & 0.882\tiny{\emph{0.089}} & 0.796\tiny{\emph{0.140}} & 0.713\tiny{\emph{0.180}} & 0.713\tiny{\emph{0.180}} & 0.879\tiny{\emph{0.119}} \\
2-\methodname              & 0.814\tiny{\emph{0.168}} & 0.760\tiny{\emph{0.186}} & 0.760\tiny{\emph{0.186}} & 0.869\tiny{\emph{0.161}} & 0.878\tiny{\emph{0.073}} & 0.827\tiny{\emph{0.102}} & 0.827\tiny{\emph{0.102}} & 0.929\tiny{\emph{0.050}}  & 0.824\tiny{\emph{0.128}} & 0.759\tiny{\emph{0.180}} & 0.759\tiny{\emph{0.180}} & 0.888\tiny{\emph{0.090}} & 0.824\tiny{\emph{0.124}} & 0.754\tiny{\emph{0.157}} & 0.754\tiny{\emph{0.157}} & 0.895\tiny{\emph{0.104}} \\ \hdashline
3-MHP             & 0.772\tiny{\emph{0.181}} & 0.789\tiny{\emph{0.194}} & 0.652\tiny{\emph{0.232}} & 0.876\tiny{\emph{0.154}} & 0.780\tiny{\emph{0.185}} & 0.796\tiny{\emph{0.217}} & 0.625\tiny{\emph{0.303}} & 0.919\tiny{\emph{0.073}}  & 0.739\tiny{\emph{0.176}} & 0.767\tiny{\emph{0.193}} & 0.562\tiny{\emph{0.289}} & 0.888\tiny{\emph{0.093}} & 0.715\tiny{\emph{0.194}} & 0.752\tiny{\emph{0.217}} & 0.523\tiny{\emph{0.297}} & 0.869\tiny{\emph{0.136}} \\
3-\methodname              & 0.804\tiny{\emph{0.169}} & 0.819\tiny{\emph{0.174}} & 0.713\tiny{\emph{0.199}} & 0.881\tiny{\emph{0.154}} & 0.885\tiny{\emph{0.082}} & 0.900\tiny{\emph{0.080}} & 0.811\tiny{\emph{0.137}} & 0.943\tiny{\emph{0.045}}  & 0.817\tiny{\emph{0.134}} & 0.835\tiny{\emph{0.136}} & 0.720\tiny{\emph{0.202}} & 0.897\tiny{\emph{0.086}} & 0.818\tiny{\emph{0.151}} & 0.837\tiny{\emph{0.149}} & 0.716\tiny{\emph{0.213}} & 0.901\tiny{\emph{0.120}} \\ \hdashline
4-MHP             & 0.773\tiny{\emph{0.170}} & 0.752\tiny{\emph{0.192}} & 0.623\tiny{\emph{0.225}} & 0.886\tiny{\emph{0.142}} & 0.830\tiny{\emph{0.131}} & 0.817\tiny{\emph{0.164}} & 0.674\tiny{\emph{0.264}} & 0.933\tiny{\emph{0.049}}  & 0.796\tiny{\emph{0.134}} & 0.783\tiny{\emph{0.157}} & 0.636\tiny{\emph{0.243}} & 0.904\tiny{\emph{0.072}} & 0.751\tiny{\emph{0.153}} & 0.726\tiny{\emph{0.189}} & 0.547\tiny{\emph{0.251}} & 0.896\tiny{\emph{0.104}} \\
4-\methodname              & 0.804\tiny{\emph{0.163}} & 0.786\tiny{\emph{0.177}} & 0.693\tiny{\emph{0.197}} & 0.889\tiny{\emph{0.147}} & 0.875\tiny{\emph{0.084}} & 0.863\tiny{\emph{0.102}} & 0.776\tiny{\emph{0.143}} & 0.945\tiny{\emph{0.042}}  & 0.812\tiny{\emph{0.135}} & 0.794\tiny{\emph{0.165}} & 0.681\tiny{\emph{0.221}} & 0.907\tiny{\emph{0.075}} & 0.786\tiny{\emph{0.152}} & 0.766\tiny{\emph{0.179}} & 0.632\tiny{\emph{0.228}} & 0.896\tiny{\emph{0.111}} \\ \hdashline
6-MHP             & 0.647\tiny{\emph{0.152}} & 0.703\tiny{\emph{0.203}} & 0.121\tiny{\emph{0.175}} & 0.886\tiny{\emph{0.131}} & 0.790\tiny{\emph{0.072}} & 0.840\tiny{\emph{0.089}} & 0.400\tiny{\emph{0.186}} & 0.939\tiny{\emph{0.038}}  & 0.749\tiny{\emph{0.116}} & 0.777\tiny{\emph{0.151}} & 0.428\tiny{\emph{0.169}} & 0.900\tiny{\emph{0.083}} & 0.649\tiny{\emph{0.132}} & 0.703\tiny{\emph{0.181}} & 0.156\tiny{\emph{0.126}} & 0.881\tiny{\emph{0.103}} \\
6-\methodname              & 0.795\tiny{\emph{0.178}} & 0.799\tiny{\emph{0.190}} & 0.648\tiny{\emph{0.234}} & 0.889\tiny{\emph{0.154}} & 0.869\tiny{\emph{0.090}} & 0.873\tiny{\emph{0.098}} & 0.745\tiny{\emph{0.167}} & 0.948\tiny{\emph{0.036}}  & 0.814\tiny{\emph{0.136}} & 0.818\tiny{\emph{0.149}} & 0.651\tiny{\emph{0.238}} & 0.911\tiny{\emph{0.070}} & 0.798\tiny{\emph{0.143}} & 0.806\tiny{\emph{0.149}} & 0.608\tiny{\emph{0.244}} & 0.906\tiny{\emph{0.106}} \\ \hdashline
8-MHP             & 0.625\tiny{\emph{0.152}} & 0.658\tiny{\emph{0.215}} & 0.099\tiny{\emph{0.140}} & 0.896\tiny{\emph{0.121}} & 0.752\tiny{\emph{0.072}} & 0.801\tiny{\emph{0.096}} & 0.260\tiny{\emph{0.148}} & 0.944\tiny{\emph{0.033}}  & 0.698\tiny{\emph{0.117}} & 0.708\tiny{\emph{0.158}} & 0.309\tiny{\emph{0.181}} & 0.908\tiny{\emph{0.069}} & 0.616\tiny{\emph{0.135}} & 0.627\tiny{\emph{0.206}} & 0.134\tiny{\emph{0.111}} & 0.897\tiny{\emph{0.092}} \\
8-\methodname              & 0.790\tiny{\emph{0.170}} & 0.798\tiny{\emph{0.185}} & 0.595\tiny{\emph{0.232}} & 0.899\tiny{\emph{0.138}} & 0.875\tiny{\emph{0.096}} & 0.878\tiny{\emph{0.113}} & 0.745\tiny{\emph{0.158}} & 0.950\tiny{\emph{0.037}}  & 0.810\tiny{\emph{0.141}} & 0.815\tiny{\emph{0.162}} & 0.632\tiny{\emph{0.229}} & 0.910\tiny{\emph{0.075}} & 0.798\tiny{\emph{0.161}} & 0.812\tiny{\emph{0.175}} & 0.586\tiny{\emph{0.241}} & 0.901\tiny{\emph{0.114}} \\ \hdashline
10-MHP            & 0.706\tiny{\emph{0.174}} & 0.745\tiny{\emph{0.204}} & 0.281\tiny{\emph{0.231}} & 0.894\tiny{\emph{0.126}} & 0.724\tiny{\emph{0.183}} & 0.761\tiny{\emph{0.222}} & 0.339\tiny{\emph{0.285}} & 0.938\tiny{\emph{0.039}}  & 0.629\tiny{\emph{0.177}} & 0.667\tiny{\emph{0.219}} & 0.181\tiny{\emph{0.210}} & 0.906\tiny{\emph{0.068}} & 0.690\tiny{\emph{0.168}} & 0.733\tiny{\emph{0.218}} & 0.223\tiny{\emph{0.196}} & 0.898\tiny{\emph{0.094}} \\
10-\methodname             & 0.793\tiny{\emph{0.173}} & 0.805\tiny{\emph{0.185}} & 0.603\tiny{\emph{0.214}} & 0.899\tiny{\emph{0.144}} & 0.866\tiny{\emph{0.101}} & 0.880\tiny{\emph{0.111}} & 0.692\tiny{\emph{0.179}} & 0.951\tiny{\emph{0.035}}  & 0.801\tiny{\emph{0.147}} & 0.813\tiny{\emph{0.166}} & 0.579\tiny{\emph{0.255}} & 0.918\tiny{\emph{0.065}} & 0.768\tiny{\emph{0.181}} & 0.791\tiny{\emph{0.196}} & 0.513\tiny{\emph{0.246}} & 0.885\tiny{\emph{0.140}} \\ \bottomrule
\end{tabular}%
}
\end{table}

\noindent \underline{{\textbf{Qualitative results}}} of \methodname on \isictestname (Fig.~\ref{fig:isic_v1_outputs} (a)) show plausibility (all \segs cover the lesion with varying degrees of over- or under-segmentation) as well as semantic consistency across \segs (e.g., when $\nstyles=3$, yellow always has a tight and jagged boundary while blue always has a loose boundary). We also provide a quantitative assessment in \suppmat Fig.~\ref{fig:suppmat_fig1} (f). 
Also, observe that in lesions with well-defined borders (top two rows), the predicted styles are similar, whereas in lesions with ambiguous borders (bottom two rows), the predictions exhibit considerable diversity. It is also worth noting that several images in \isictestname have either incorrect or imprecise \say{\groundtruth{}} masks (Fig.~\ref{fig:isic_v1_outputs} (b)),
which leads to incorrect penalization of \methodname's accurate predictions during evaluation.

\noindent \underline{{\textbf{Quantitative results:}}} 
For \ssegname methods, we report the Dice coefficient.
For \msegname methods, we report ${\texttt{max}}_j(d)$, where $d = \mathrm{Dice}(Y_{ik}, \hat{Y}_{ij})$, to assess the highest agreement, and $\{{\texttt{mean}}_j(d), \ {\texttt{median}}_j(d), \ {\texttt{min}}_j(d)\}$ to assess the plausibility of all $\nstyles$ \segs.
For example, an \msegname model that produces even one poor \seg will have low scores for ${\texttt{min}}_j(d)$, indicating low plausibility.

\begin{table}[ht!]
\centering
\caption{\methodname's \seg agreement (mean\tiny{\emph{std.dev.}}\normalsize \ of $\mathrm{Dice}_{\mathrm{IASS}}$ and $\mathrm{Dice}_{\mathrm{ASSS}}$) \normalsize and style alignment (\newmetricnameshort) on the 27 annotator preferences in \newdatasetname.
\camready{$\beststyle$ denotes the single style that, for each row, maximizes agreement with the ground truth.}
As more styles are modeled, $\mathrm{Dice}_{\mathrm{IASS}}$, $\mathrm{Dice}_{\mathrm{ASSS}}$, and \newmetricnameshort all improve, and all annotator preferences consistently align with a discovered style.}
\label{tab:v2_results}
\resizebox{\textwidth}{!}{%
\setlength{\tabcolsep}{1.35em}
\def\arraystretch{1.25}
\begin{tabular}{@{}cc|c|c:cc|c:cc|c:cc@{}}
\toprule
\multirow{2}{*}{\begin{tabular}[c]{@{}c@{}}Annotator + Tool\\  + Experience\end{tabular}} & \multirow{2}{*}{\begin{tabular}[c]{@{}c@{}}Seg.\\Count\end{tabular}} & 1-\methodname & \multicolumn{3}{c|}{2-\methodname} & \multicolumn{3}{c|}{3-\methodname} & \multicolumn{3}{c}{4-\methodname} \\ \cmidrule(l){3-12} 
 
 &  & $\mathrm{Dice}_{\mathrm{ISSS}}$ & $\mathrm{Dice}_{\mathrm{ISSS}}$ & $\mathrm{Dice}_{\mathrm{ASSS}}$ & \begin{tabular}[c|]{@{}c@{}}\camready{$\beststyle$}\end{tabular} & $\mathrm{Dice}_{\mathrm{ISSS}}$ & $\mathrm{Dice}_{\mathrm{ASSS}}$ & \begin{tabular}[c|]{@{}c@{}}\camready{$\beststyle$}\end{tabular} & $\mathrm{Dice}_{\mathrm{ISSS}}$ & $\mathrm{Dice}_{\mathrm{ASSS}}$ & \camready{$\beststyle$} \\ 
 
 \midrule
A00+T2+E & 1573 & 0.892\tiny{\emph{0.089}} & 0.923\tiny{\emph{0.061}} & 0.913\tiny{\emph{0.087}} & 2 & 0.944\tiny{\emph{0.049}} & 0.913\tiny{\emph{0.106}} & 3 & 0.944\tiny{\emph{0.044}} & 0.914\tiny{\emph{0.111}} & 1 \\
A00+T2+N & 1305 & 0.716\tiny{\emph{0.302}} & 0.761\tiny{\emph{0.293}} & 0.728\tiny{\emph{0.308}} & 2 & 0.793\tiny{\emph{0.287}} & 0.727\tiny{\emph{0.313}} & 3 & 0.790\tiny{\emph{0.290}} & 0.726\tiny{\emph{0.304}} & 3 \\
\hdashline
A01+T1+N & 6 & 0.559\tiny{\emph{0.362}} & 0.766\tiny{\emph{0.152}} & 0.766\tiny{\emph{0.152}} & 1 & 0.754\tiny{\emph{0.132}} & 0.741\tiny{\emph{0.125}} & 2 & 0.819\tiny{\emph{0.106}} & 0.767\tiny{\emph{0.113}} & 2 \\
A01+T3+E & 297 & 0.900\tiny{\emph{0.104}} & 0.915\tiny{\emph{0.093}} & 0.897\tiny{\emph{0.107}} & 2 & 0.927\tiny{\emph{0.075}} & 0.900\tiny{\emph{0.097}} & 1 & 0.931\tiny{\emph{0.067}} & 0.904\tiny{\emph{0.090}} & 3 \\
A01+T3+N & 2148 & 0.829\tiny{\emph{0.185}} & 0.857\tiny{\emph{0.167}} & 0.817\tiny{\emph{0.170}} & 1 & 0.869\tiny{\emph{0.159}} & 0.836\tiny{\emph{0.178}} & 1 & 0.876\tiny{\emph{0.148}} & 0.836\tiny{\emph{0.175}} & 3 \\
\hdashline
A02+T1+E & 1742 & 0.844\tiny{\emph{0.177}} & 0.880\tiny{\emph{0.140}} & 0.856\tiny{\emph{0.159}} & 1 & 0.886\tiny{\emph{0.132}} & 0.854\tiny{\emph{0.159}} & 1 & 0.895\tiny{\emph{0.112}} & 0.859\tiny{\emph{0.148}} & 4 \\
A02+T3+E & 468 & 0.856\tiny{\emph{0.172}} & 0.889\tiny{\emph{0.167}} & 0.883\tiny{\emph{0.175}} & 2 & 0.899\tiny{\emph{0.161}} & 0.874\tiny{\emph{0.188}} & 3 & 0.903\tiny{\emph{0.146}} & 0.890\tiny{\emph{0.160}} & 1 \\
\hdashline
A03+T1+E & 1622 & 0.778\tiny{\emph{0.168}} & 0.845\tiny{\emph{0.117}} & 0.827\tiny{\emph{0.137}} & 1 & 0.854\tiny{\emph{0.111}} & 0.824\tiny{\emph{0.145}} & 2 & 0.881\tiny{\emph{0.095}} & 0.823\tiny{\emph{0.132}} & 4 \\
A03+T3+E & 260 & 0.891\tiny{\emph{0.116}} & 0.912\tiny{\emph{0.086}} & 0.876\tiny{\emph{0.173}} & 2 & 0.923\tiny{\emph{0.089}} & 0.868\tiny{\emph{0.150}} & 1 & 0.932\tiny{\emph{0.074}} & 0.874\tiny{\emph{0.163}} & 3 \\
\hdashline
A04+T1+E & 992 & 0.850\tiny{\emph{0.158}} & 0.880\tiny{\emph{0.131}} & 0.860\tiny{\emph{0.149}} & 1 & 0.888\tiny{\emph{0.132}} & 0.866\tiny{\emph{0.153}} & 2 & 0.906\tiny{\emph{0.108}} & 0.856\tiny{\emph{0.157}} & 4 \\
A04+T1+N & 61 & 0.760\tiny{\emph{0.242}} & 0.840\tiny{\emph{0.152}} & 0.823\tiny{\emph{0.164}} & 1 & 0.837\tiny{\emph{0.162}} & 0.786\tiny{\emph{0.201}} & 1 & 0.827\tiny{\emph{0.206}} & 0.789\tiny{\emph{0.226}} & 4 \\
A04+T3+E & 913 & 0.912\tiny{\emph{0.088}} & 0.939\tiny{\emph{0.054}} & 0.934\tiny{\emph{0.065}} & 2 & 0.948\tiny{\emph{0.047}} & 0.926\tiny{\emph{0.069}} & 1 & 0.951\tiny{\emph{0.045}} & 0.932\tiny{\emph{0.063}} & 3 \\
A04+T3+N & 90 & 0.877\tiny{\emph{0.096}} & 0.910\tiny{\emph{0.068}} & 0.905\tiny{\emph{0.070}} & 2 & 0.928\tiny{\emph{0.031}} & 0.908\tiny{\emph{0.044}} & 3 & 0.926\tiny{\emph{0.052}} & 0.913\tiny{\emph{0.055}} & 1 \\
\hdashline
A05+T1+E & 752 & 0.815\tiny{\emph{0.203}} & 0.862\tiny{\emph{0.163}} & 0.837\tiny{\emph{0.179}} & 1 & 0.873\tiny{\emph{0.162}} & 0.827\tiny{\emph{0.184}} & 1 & 0.882\tiny{\emph{0.147}} & 0.841\tiny{\emph{0.177}} & 4 \\
A05+T3+E & 742 & 0.875\tiny{\emph{0.129}} & 0.903\tiny{\emph{0.109}} & 0.891\tiny{\emph{0.113}} & 2 & 0.916\tiny{\emph{0.098}} & 0.878\tiny{\emph{0.120}} & 1 & 0.919\tiny{\emph{0.091}} & 0.891\tiny{\emph{0.108}} & 1 \\
\hdashline
A06+T1+E & 10 & 0.824\tiny{\emph{0.187}} & 0.902\tiny{\emph{0.037}} & 0.885\tiny{\emph{0.070}} & 1 & 0.909\tiny{\emph{0.034}} & 0.889\tiny{\emph{0.049}} & 2 & 0.909\tiny{\emph{0.039}} & 0.880\tiny{\emph{0.063}} & 4 \\
A06+T3+E & 24 & 0.862\tiny{\emph{0.079}} & 0.916\tiny{\emph{0.053}} & 0.916\tiny{\emph{0.053}} & 2 & 0.934\tiny{\emph{0.031}} & 0.923\tiny{\emph{0.031}} & 3 & 0.933\tiny{\emph{0.041}} & 0.929\tiny{\emph{0.040}} & 1 \\
\hdashline
A07+T1+E & 67 & 0.820\tiny{\emph{0.157}} & 0.877\tiny{\emph{0.124}} & 0.867\tiny{\emph{0.150}} & 1 & 0.890\tiny{\emph{0.108}} & 0.862\tiny{\emph{0.157}} & 2 & 0.897\tiny{\emph{0.104}} & 0.862\tiny{\emph{0.149}} & 4 \\
A07+T1+N & 251 & 0.837\tiny{\emph{0.141}} & 0.892\tiny{\emph{0.085}} & 0.879\tiny{\emph{0.104}} & 1 & 0.903\tiny{\emph{0.067}} & 0.875\tiny{\emph{0.114}} & 2 & 0.905\tiny{\emph{0.070}} & 0.873\tiny{\emph{0.101}} & 4 \\
A07+T3+E & 12 & 0.925\tiny{\emph{0.055}} & 0.938\tiny{\emph{0.019}} & 0.937\tiny{\emph{0.019}} & 2 & 0.939\tiny{\emph{0.020}} & 0.916\tiny{\emph{0.055}} & 1 & 0.947\tiny{\emph{0.016}} & 0.932\tiny{\emph{0.017}} & 1 \\
A07+T3+N & 39 & 0.863\tiny{\emph{0.177}} & 0.918\tiny{\emph{0.061}} & 0.913\tiny{\emph{0.071}} & 2 & 0.933\tiny{\emph{0.037}} & 0.899\tiny{\emph{0.148}} & 3 & 0.934\tiny{\emph{0.039}} & 0.914\tiny{\emph{0.079}} & 1 \\
\hdashline
A08+T1+E & 26 & 0.666\tiny{\emph{0.225}} & 0.750\tiny{\emph{0.161}} & 0.680\tiny{\emph{0.242}} & 2 & 0.747\tiny{\emph{0.197}} & 0.653\tiny{\emph{0.260}} & 1 & 0.793\tiny{\emph{0.134}} & 0.666\tiny{\emph{0.261}} & 1 \\
A08+T3+E & 111 & 0.605\tiny{\emph{0.230}} & 0.668\tiny{\emph{0.197}} & 0.626\tiny{\emph{0.210}} & 1 & 0.677\tiny{\emph{0.206}} & 0.628\tiny{\emph{0.218}} & 2 & 0.735\tiny{\emph{0.166}} & 0.669\tiny{\emph{0.203}} & 2 \\
\hdashline
A09+T1+E & 30 & 0.815\tiny{\emph{0.121}} & 0.841\tiny{\emph{0.098}} & 0.784\tiny{\emph{0.156}} & 1 & 0.873\tiny{\emph{0.089}} & 0.833\tiny{\emph{0.113}} & 2 & 0.884\tiny{\emph{0.076}} & 0.812\tiny{\emph{0.119}} & 4 \\
A09+T1+N & 1 & 0.953\tiny{\emph{0.000}} & 0.927\tiny{\emph{0.000}} & 0.927\tiny{\emph{0.000}} & 2 & 0.955\tiny{\emph{0.000}} & 0.955\tiny{\emph{0.000}} & 1 & 0.947\tiny{\emph{0.000}} & 0.947\tiny{\emph{0.000}} & 3 \\
A09+T3+E & 10 & 0.900\tiny{\emph{0.074}} & 0.918\tiny{\emph{0.054}} & 0.918\tiny{\emph{0.054}} & 2 & 0.933\tiny{\emph{0.038}} & 0.909\tiny{\emph{0.044}} & 1 & 0.937\tiny{\emph{0.043}} & 0.919\tiny{\emph{0.040}} & 3 \\
A09+T3+N & 3 & 0.894\tiny{\emph{0.070}} & 0.911\tiny{\emph{0.058}} & 0.911\tiny{\emph{0.058}} & 2 & 0.957\tiny{\emph{0.015}} & 0.957\tiny{\emph{0.015}} & 3 & 0.944\tiny{\emph{0.030}} & 0.944\tiny{\emph{0.030}} & 1 \\ 
\midrule
\multicolumn{2}{c|}{\newmetricnameshort (Eqn.~\ref{eqn:newmetricnameshort})} & - & \multicolumn{3}{c|}{0.299\tiny{\emph{0.208}}} & \multicolumn{3}{c|}{0.347\tiny{\emph{0.237}}} & \multicolumn{3}{c}{0.466\tiny{\emph{0.296}}} \\
\bottomrule
\end{tabular}%
}
\end{table}

Table~\ref{tab:v1_results} shows that predicting more than one style (\methodname, MHP) improves performance (${\texttt{max}}_j(d)$) compared to \ssegname methods, and even \msegname methods that predict just two styles ($2$-\methodname, $2$-MHP) consistently outperform \ssegname methods. 
Moreover, as $\nstyles$ increases, a larger number of diverse \segs are produced, and the ${\texttt{max}}_j(d)$ keeps improving. However, we observe that for three out of the four datasets, the ${\texttt{max}}_j(d)$ performance either plateaus or starts to decline as $\nstyles$ increases. We posit that after an optimal number of styles, generating more \segs leads to diversity at the cost of performance,
\camready{and leave this investigation for future work.} 
Interestingly, datasets that do not have a documented presence of inter-\seg variability (\dermofit, \phtwo, SCD) also benefit from learning to predict multiple \segs, indicating style variability in the \groundtruth masks. 
A post hoc investigation of \dermofit, for example, confirms the presence of different annotation styles (difference in boundary granularity; \suppmat Fig.~\ref{fig:suppmat_fig1} (b)).

Finally, \methodname consistently outperforms MHP for all datasets and $\nstyles$, except $\nstyles=10$ with SCD, and 
as $\nstyles$ increases, the plausibility of MHP models across all the predictions decreases, as evident through the declining $\{{\texttt{mean}}_j(d),$ ${\texttt{median}}_j(d), {\texttt{min}}_j(d)\}$ scores. For example, when modeling 10 styles, the [${\texttt{min}}_j(d)$, ${\texttt{max}}_j(d)$] range across 10 segmentations for 10,000 test images in \isictestname is [0.281, 0.894] for 10-MHP and [0.603, 0.899] for 10-\methodname, meaning all the predicted \segs are more plausible for the latter. We attribute this improvement to 
the plausibility constraint ($\mathcal{L}_2$ in Eqn.~\ref{eqn:l2}), which penalizes predicted \segs that considerably deviate from the \groundtruth.

\noindent \underline{\textbf{A new multi-annotator dataset:}} Next, we propose {\newdatasetname}, a new multi-annotator SLS dataset curated from the ISIC Archive that, to the best of our knowledge, is the largest such dataset to contain annotator correspondence. The annotator-\seg mapping in \newdatasetname forms an incomplete bipartite graph, i.e., not every image has been segmented by every annotator. \newdatasetname contains 12,951 images segmented by 10 annotators, resulting in 13,555 image-mask pairs (breakdown in Fig.~\ref{fig:datasets_overview} (c)). Unlike other multi-annotator 
datasets, the variability in \newdatasetname's \segs stems from three annotation pipeline factors: the annotator (10 annotator IDs: \say{A00}--\say{A09}), the tool used (\say{T1}, \say{T2}, \say{T3}), and the expertise of the manual reviewer (\say{expert} or \say{novice})~\cite{mirikharaji2021d},
resulting in 27 unique annotator preferences, which we use for our evaluation.
We measure \methodname performance in two settings: (i) \textbf{image-adaptive style selection (IASS)}: for every image, we find the style that maximizes the agreement with \groundtruth, measured as $\mathrm{Dice}_{\mathrm{IASS}} = {\texttt{max}}_j(\mathrm{Dice}(Y_{ik}, \hat{Y}_{ij}))$, and (ii) a more challenging \textbf{annotator-specific style selection (ASSS)}: we find a single style, fixed across all images, that maximizes agreement with \groundtruth, measured as $\mathrm{Dice}_{\mathrm{ASSS}} = \mathrm{Dice}(Y_{ik}, \hat{Y}_{i\camready{\beststyle}})$, where $\camready{\beststyle} = \argmax_j (\sum_i \mathrm{Dice}(Y_{ik}, \hat{Y}_{ij}))$. Note that $\mathrm{Dice}_{\mathrm{ASSS}} \leq \mathrm{Dice}_{\mathrm{IASS}}$.

Quantitative results of \methodname on \newdatasetname (Table~\ref{tab:v2_results}; \camready{additional results in \suppmat Fig.~\ref{fig:suppmat_fig1} (c)})
% and \suppmat Fig.~\ref{fig:suppmat_fig1} (c)) 
show that each of the discovered styles presents a high agreement with almost all the annotator preferences. 
A notable outlier is \say{A08}, and upon manual inspection, we found 
a large number of \groundtruth \segs to be incorrect to varying degrees (\suppmat Fig.~\ref{fig:suppmat_fig1} (e)),
which explains the lower evaluation performance. 
Similar to Table~\ref{tab:v1_results}, modeling even two styles yields better performance than one style.
Moreover, as $\nstyles$ increases, the newly discovered styles continue to show increasing usefulness, since all of them consistently align with one or more annotator preferences, meaning that they are able to capture the diversity in \segs with an increasing level of granularity. 

\camready{As an additional experiment to assess whether the learned styles are able to model tool-specific (\say{T1}, \say{T2}, \say{T3}) latent factors, 
we separate the segmentations into three groups based on the tool, pass each corresponding image through a trained 3-StyleSeg model, and determine the predicted segmentation style with the highest overlap. We observe that the most commonly chosen style within a group is unique for each of the three tool groups, suggesting that differences among the three tools are learned within the three styles.
}

% JER SUGGESTION. 

\noindent \underline{\textbf{A new style alignment measure:}}
When choosing only one style to evaluate each annotator preference, it is important to note that a particular style could be assigned as the chosen style for a certain annotator even if it best fits either 100\% (perfect alignment) of the images or just slightly above random chance (weak alignment). We propose to measure this \newmetricname (\newmetricnameshort) as 1 -- `normalized Shannon entropy of annotator-style assignment', i.e., 
\begin{equation}
\label{eqn:newmetricnameshort}
    \mathrm{AS}^2 = 1 - \frac{-\sum_{i=1}^{\nstyles} q_i \log_2 q_i}{-\sum_{j=1}^{\nstyles} \frac{1}{\nstyles} \log_2 \frac{1}{\nstyles}},
\end{equation}
where $q_i$ is the vector of fractions of \segs assigned to each style (e.g., $q = [0.70, 0.15, 0.15]$ for a $3$-\methodname model that assigns 100 images from a certain annotator preference as 70:15:15 images for styles 1, 2, and 3, yielding \newmetricnameshort = $0.255$).
Note that \newmetricnameshort is 0 for uniform assignments, and increases logarithmically approaching 1 as assignments become more consistent.
Our results in Table~\ref{tab:v2_results} show that \newmetricnameshort values do not decrease as $\nstyles$ increases, meaning that learning to model more styles is not detrimental to \seg quality and indeed captures more diversity. 
Additional results are presented in \suppmat Fig.~\ref{fig:suppmat_fig1} (d, g).

\section{Conclusion}
\label{sec:conclusion}

We formulated the problem of segmentation style discovery in the absence of annotator correspondence. 
We showed how our proposed method, \methodname, discovers segmentation styles that are diverse, semantically consistent, and more plausible than those generated by competing methods, as evaluated on four public skin lesion \seg (SLS) datasets. We also curated
\newdatasetname, the largest multi-annotator SLS dataset with 13,555 image-mask pairs from 10 annotators from ISIC Archive, and showed how \methodname consistently achieves high agreement with the annotator preferences, as measured through the Dice coefficient as well as a newly proposed measure, \newmetricname, for measuring annotator-style alignment.
Future work would include an explicit \say{disentanglement} of annotation styles from image content and approaches to find the optimal number of styles in a \seg dataset.

\begin{credits}
\subsubsection{\ackname} 
The authors are grateful for the computational resources provided by NVIDIA Corporation and Digital Research Alliance of Canada (formerly Compute Canada). Partial funding for this project was provided by the Natural Sciences and Engineering Research Council of Canada (NSERC RGPIN/06752-2020).

\subsubsection{\discintname}
The authors have no competing interests to declare.
\end{credits}
\bibliographystyle{splncs04}
\bibliography{references}

\clearpage
\pagenumbering{arabic}
\setcounter{table}{0}
\setcounter{figure}{0}
\renewcommand{\thetable}{SM\arabic{table}}
\renewcommand{\thefigure}{SM\arabic{figure}}
\appendix

\section*{Supplementary Material}

\subsection*{Implementation Details}

The style classification model $\classmodel: (X_i, Y_{ik})\in \mathbb{R}^{224 \times 224 \times 4} \to p_i \in \mathbb{R}^{M}$ is an ImageNet-pretrained ResNet-50 model with two modifications: it takes 4-channels (i.e., a concatenation of $X_i$ and $Y_{ik}$) as input and produces a $\nstyles$-class prediction. The \seg model $\segmodel : X_i \in \mathbb{R}^{224 \times 224 \times 3} \to \hat{Y_i} \in \mathbb{R}^{224 \times 224 \times M}$ is an ImageNet-pretrained VGG-16 model with the fully connected layers removed, and multi-scale features resized, concatenated, and passed through a Conv2D layer with a sigmoid activation for binary mask prediction~\cite{kawahara2018fully}.
The images and masks were resized to $224 \times 224$ spatial resolution using nearest-neighbor interpolation. All models were trained for 10 epochs with a batch size of 4 using the Adam optimizer and a learning rate of 5e-5, and the epoch with the lowest loss $\mathcal{L}_\mathrm{total}$ (Eqn.~\ref{eqn:total_loss}) on the validation set was used for evaluation. All models were trained on an Ubuntu 20.04 workstation with AMD Ryzen 9 5950X, 32 GB of RAM, and NVIDIA RTX 3090 GPU, running Python 3.10.13 and PyTorch 2.1.2. The PyTorch implementation of \methodname 
% will be made publicly available on GitHub upon acceptance.
and more details about \newdatasetname are available at {\url{\ghrepo}}.

% \subsection*{\newdatasetname Dataset Details}

% \input{figures/suppmat_figure1}
% \begin{figure}[ht!]
% \centering
% \begin{minipage}{.66\textwidth}
%   \centering
%   \includegraphics[width=\textwidth]{figs/V1Train_Metrics_Distribution.pdf}
%   \captionof{figure}{Distribution of image-level pairwise Dice coefficient and Fleiss' kappa values in the 4704 training image-mask pairs from ISIC Archive.}
%   \label{fig:test1}
% \end{minipage}
% \hfill
% \begin{minipage}{.3\textwidth}
%   \centering
%   \includegraphics[width=\textwidth]{figs/DermoFit_Segs.pdf}
%   \captionof{figure}{Undocumented annotation style variability in \dermofit~\cite{ballerini2013color} masks.}
%   \label{fig:test2}
% \end{minipage}
% \end{figure}

\begin{figure*}[ht!]
    \centering
    \begin{subfigure}[t]{0.66\textwidth}
        \centering
        \includegraphics[width=\textwidth]{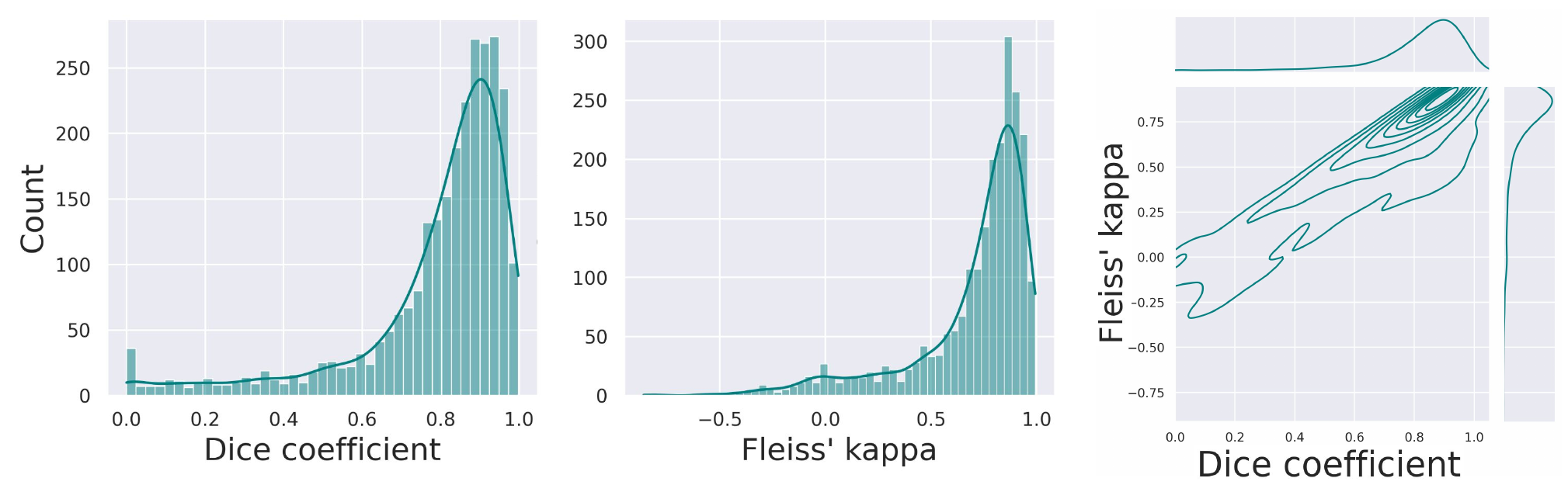}
        \caption{Distribution of image-level pairwise Dice coefficient and Fleiss' kappa values in the 4,704 training image-mask pairs from ISIC Archive. Note how a considerable number of images have poor inter-annotator agreement (Dice $< 0.2$ and Fleiss' kappa $< 0$).}
    \end{subfigure}
    \hfill
    \begin{subfigure}[t]{0.3\textwidth}
        \centering
        \includegraphics[width=\textwidth]{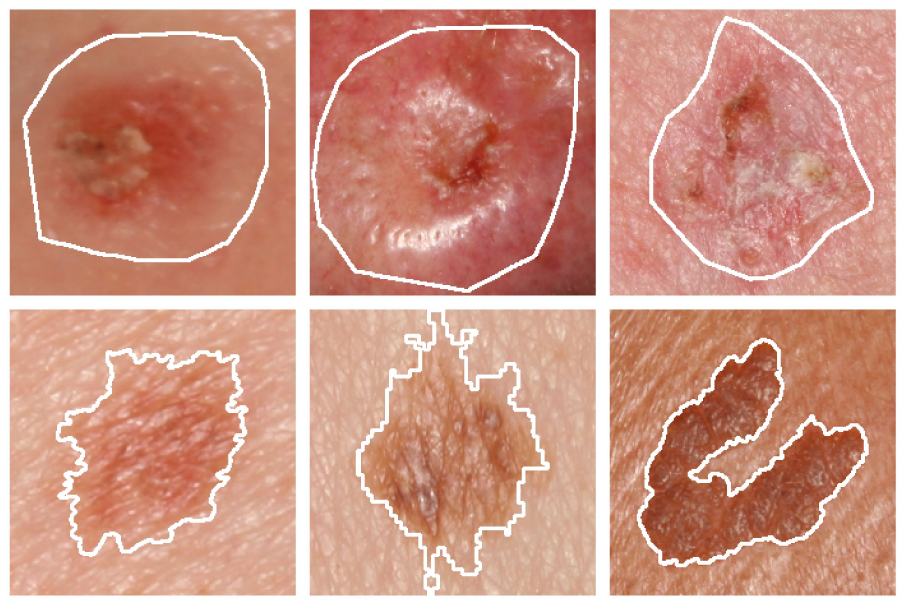}
        \caption{Undocumented annotation style variability (varying boundary granularity) in \dermofit's~\cite{ballerini2013color} \groundtruth masks.}
    \end{subfigure}
    \\
    \begin{subfigure}[t]{0.65\textwidth}
        \centering
        \includegraphics[width=\textwidth]{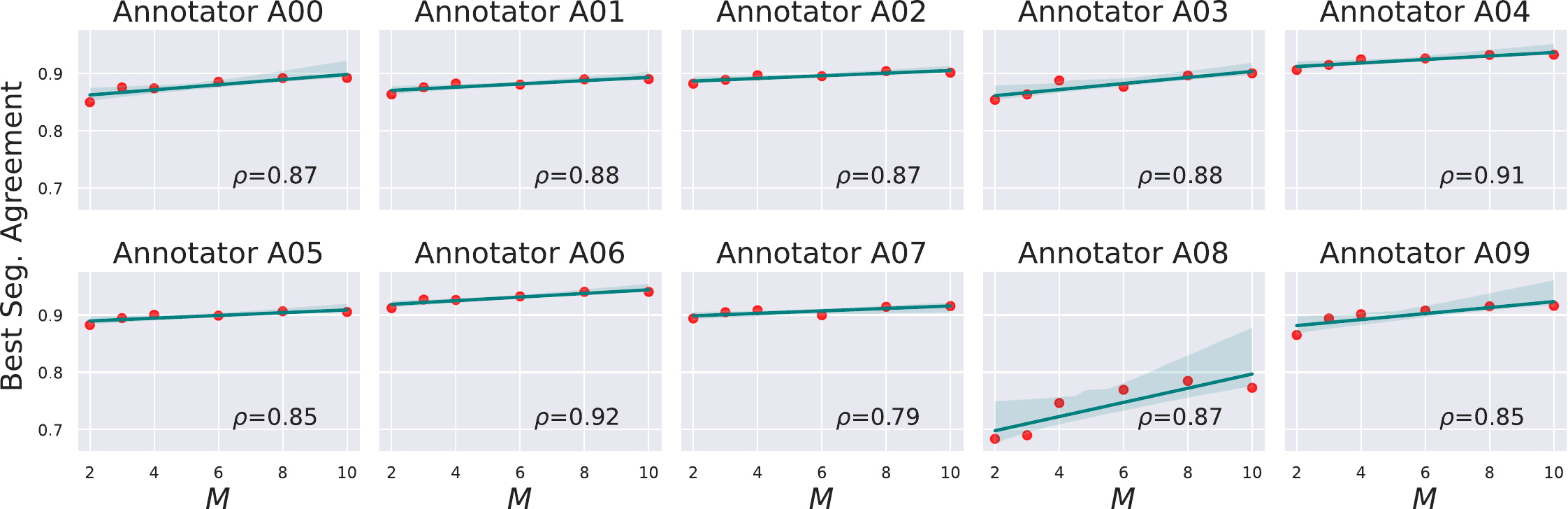}
        \caption{\methodname \seg performance (Dice) on \newdatasetname, reported at annotator-level (\say{A00}--\say{A09}). Note how the Dice values are strongly correlated (Pearson correlation coefficient $\rho$) with the number of styles ($M$).}
    \end{subfigure}%
    \hfill
    \begin{subfigure}[t]{0.3\textwidth}
        \centering
        \includegraphics[width=\textwidth]{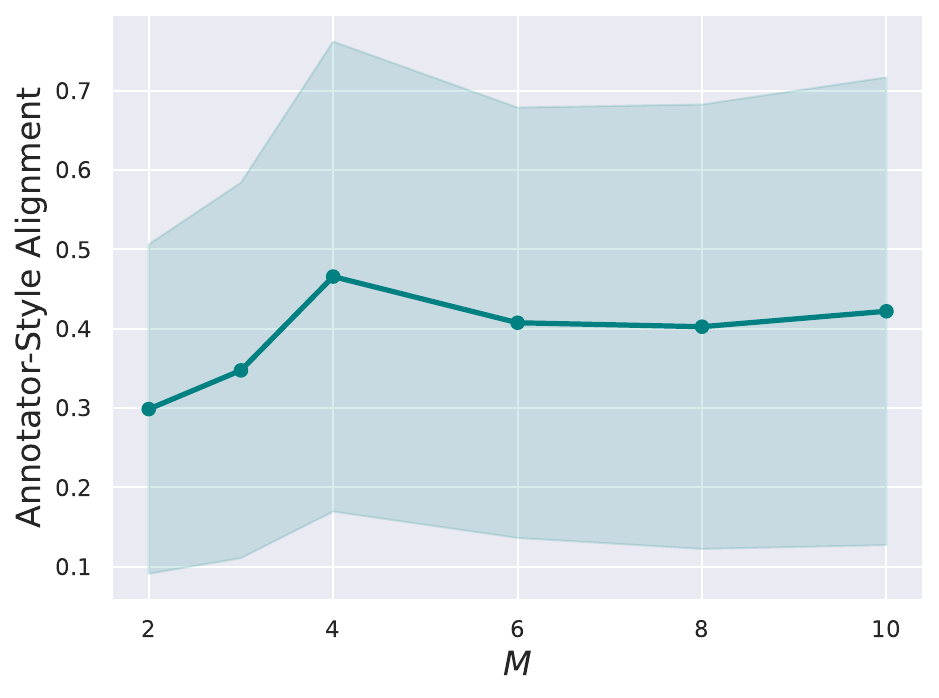}
        \caption{\newmetricnameshort values remain high as $\nstyles$ increases, meaning modeling more styles captures larger diversity.}
    \end{subfigure}
    \caption{Supplementary Figures}
    \label{fig:suppmat_fig1}

\end{figure*}
\begin{figure}[ht!]\ContinuedFloat

    \begin{subfigure}[t]{\textwidth}
        \centering
        \includegraphics[width=\textwidth]{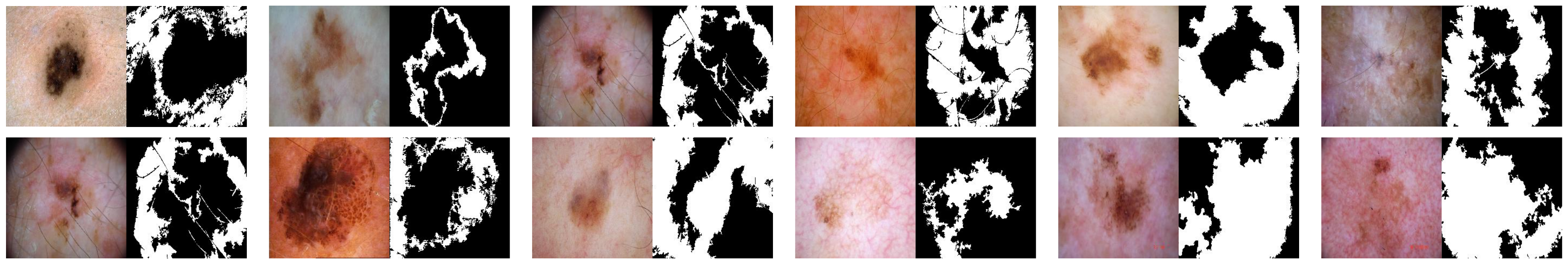}
        \caption{Images and \say{\groundtruth{}} \seg masks from annotator \say{A08} in \newdatasetname. Note the poor quality of \seg, which in turn, affects evaluation.}
    \end{subfigure}
    \\
    \begin{subfigure}[t]{\textwidth}
        \centering
        \includegraphics[width=\textwidth]{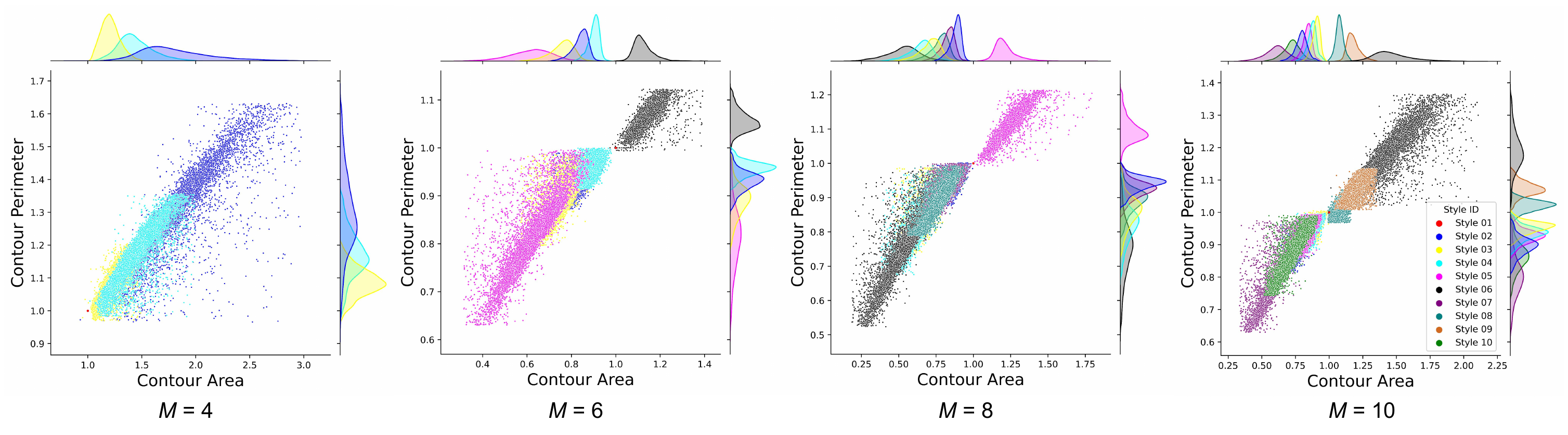}
        \caption{Evaluating semantic consistency of \seg styles on \isictestname. For all the 10,000 images, the shape features (area and perimeter of \seg contours) of \methodname outputs are calculated, normalized per lesion w.r.t. the first style (red dot at (1.0, 1.0)), and the points and their kernel density estimates are colored by their style. Even for large values of $\nstyles$, the styles remain distinct. Note that the styles vary in their contour area indicating under- and over-segmentation of skin lesions. Also,
        for a certain contour area ($A$), the contour perimeter ($P$) varies, implying the styles differ in higher-order features such as border irregularity index and contour compactness, which is expected since both border irregularity $=\frac{P^2}{4\pi A}$~\cite{ercal1994neural} and compactness $=\frac{\mathrm{convex \ hull \ area}}{\mathrm{A}}$ are also functions of perimeter and area.
        % We choose to study contour area and perimeter since higher-order features such as border irregularity and compactness are also functions of perimeter and area.
        }
    \end{subfigure}
    \\
    \vspace{10mm}
    \begin{subfigure}[t]{\textwidth}
        \centering
        \includegraphics[width=\textwidth]{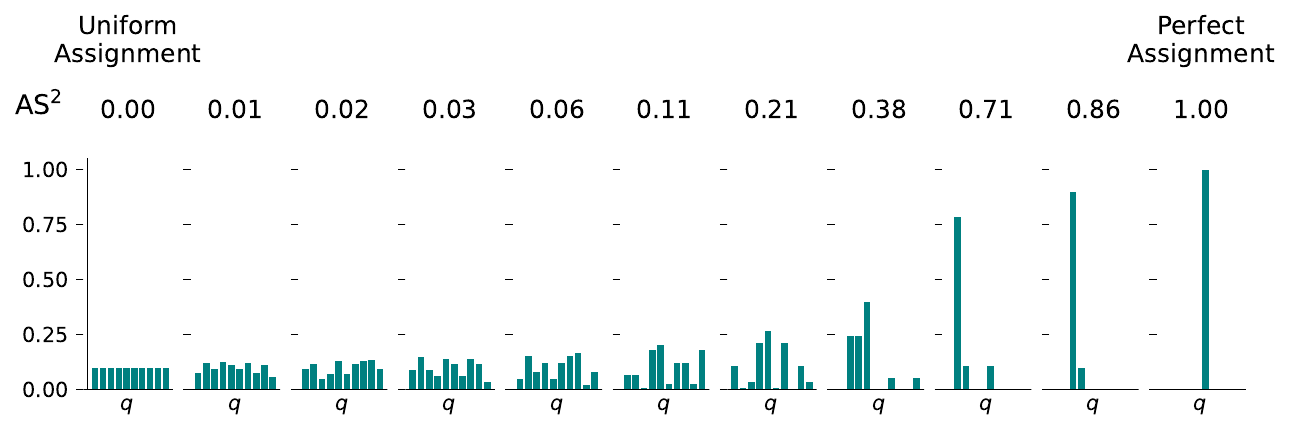}
        \caption{\newmetricname (\newmetricnameshort) values for a variety of fractions of segmentations $q$ (Eqn.~\ref{eqn:newmetricnameshort}) with $\nstyles=10$. \newmetricnameshort is 1 in case of a perfect assignment and 0 in case of a uniform assignment. \newmetricnameshort increases logarithmically as $q$ becomes more concentrated: for example, note that even for a low-entropy assignment such as $[0.9, 0.1, 0.0, \cdots 0.0]$, the \newmetricname drops to $0.86$.
        }
    \end{subfigure}

\caption{
\textbf{(continued)} Supplementary Figures.
}
\end{figure}

% ✓ ASA line plot for all M
% ✓ V2b-granular regression line plot
% ✓ DermoFit border variability samples
% ✓ Distribution of Dice and Fleiss' kappa.
% ✓ Area-perimeter relation plot
% (...) Tool-level (T1, T2, T3) classification accuracy with 3 styles.
% ✗ Distribution of training images.

%
\end{document}